\documentclass[conference]{IEEEtran}
\IEEEoverridecommandlockouts
\usepackage{amsmath,amssymb,amsfonts}
\usepackage{algorithmic}
\usepackage{graphicx}
 \graphicspath{{/Images/}}
\usepackage{textcomp}
\usepackage{xcolor}
\usepackage{float}

\usepackage[style=ieee, sorting=none]{biblatex}
\bibliography{JJ_vdv_Research_paper}
\def\BibTeX{{\rm B\kern-.05em{\sc i\kern-.025em b}\kern-.08em
    T\kern-.1667em\lower.7ex\hbox{E}\kern-.125emX}}
\begin{document}
\title{A Deep Neural Network Approach to Fare Evasion\\
{\footnotesize \textsuperscript{*}Note: This is a research project regarding Computer Vision, to detect fare evasions.}
}

\author{\IEEEauthorblockN{Johannes van der Vyver}
\IEEEauthorblockA{\textit{Faculty of Computer Science} \\
\textit{University of Johannesburg}\\
Johannesburg, South Africa}
}

\maketitle

\begin{abstract}
Fare evasion is a problem for public transport companies, with LSTM models this issue can help companies get an analytical insight into where this issue occurs the most, to prevent capital loss. In addition to the financial burden this problem causes, having more inspectors is not enough to alleviate the problem. The purpose of this study is to find a different way to predict fare evasion in the public transport sector. Through the use of keypoint extractions of passengers in video footage, an LSTM model is trained on those keypoints to help predict the actions of passengers between payments and evasions. The results were promising when it came to predicting the actions of passengers on real-time footage. Thus a sophisticated approach can help to decrease the fare evasion problem. A ReID model can be used alongside the LSTM model for better accuracy, as there is always the chance that a person might only pay for the fare at a later stage. With both models, it is possible for public transport companies to start narrowing down where the root of their fare evasion problems emerges.
\end{abstract}


\section{Introduction}
Fare evasion is a major contributor to significant financial loss in the public transport sector of South Africa, particularly bus transport. People would sneak behind someone who has paid to avoid payment themselves, through overcrowded bus entrances.

Studies into fare evasion have shown that the Chile bus service has measured approximately 28 percent of fare evasion. Increasing the inspections by ten percent, only lowered fare evasions by 0.8 percent \cite{FareEvasionTimeSeries}.

The problems regarding fare evasion for public transport in South Africa have not been completely resolved. The human factor isn't an adequate solution to the problem, as we have seen additional inspections only decrease fare evasions by a small amount.

In this paper, it will be argued that a more sophisticated approach is required to deal with this ongoing problem. The human factor lacks significant change to the fare evasion problem, thus a more robust and accurate approach is required. The specific objective of this study was to investigate a way to accurately detect passengers who commit fare evasion in the public transport sector. This study provides new insights into ways to battle this problematic issue.

\vspace{12pt}
\section{Experimental Design}
To test the primary hypothesis of using a Character action detection model with a Long-Short Term Memory (LSTM) model, we conduct an evaluation of the model's accuracy. The model's accuracy will be assessed by its Confusion Matrix. True Positive Rate TRP = TP (True positive) / (TP + FN (False Negative)). It is observed that a Convolution Neural Network (CNN) with an LSTM model would yield the best accuracy \cite{PerformanceCNN}. For this project, the mediapipe holistic for the character action detection model that is an LSTM model is used, as this will require less data and is much faster to train.\\

\subsection{Datasets}
For the first model, datasets of two types of videos -  people paying and people committing fare evasion - will be used. In using this type of data set keypoints can be extracted with the mediapipe library and the data can subsequently be labelled accordingly.\\ 

Each frame for three seconds will be saved in multiple arrays. The numpy arrays will be divided into the payment and evasion categories, which will in turn be used to train the LSTM Neural Network.\\

\subsection{Experimental Setup}
Preprocessing will have to be done on the dataset and from there we will train the character Action Recognition (CAR) model on the behaviour of the images/videos. If the results of the LSTM model after training does not have a high accuracy, the hyper-parameters will be changed and the training process will restart.\\

After training the models, it will be tested and evaluated on the test set. From there the model will be tested with video footage to see how the model performs.\\

Using a CNN-LSTM model that can predict hand signs or facial recognition can be seen as an example of the same kind of model that will created in this paper. Models like a CNN-LSTM that anticipate a driver's conduct \cite{PerformanceCNN}, is a more sophisticated model that can show the reactions of drivers in real-time traffic.\\

The experiment was repeated five times on different hyperparameters, by changing the activation function and the output layer's activation function of the Neural Network. Refining the train test split also made a difference in the performance, starting off at a 60 - 40 split respectively and leaving it at a 70 - 30 split. The use of the Adam optimizer for the Deep Neural Network was satisfactory, as seen in Figure 1. In the final evaluation of the model, the ReLU function was used as the activation function on all the layers except the output layer, where the softmax function was used.\\

\begin{figure}[H]
    \centering
    \includegraphics[width=9cm,height=5cm]{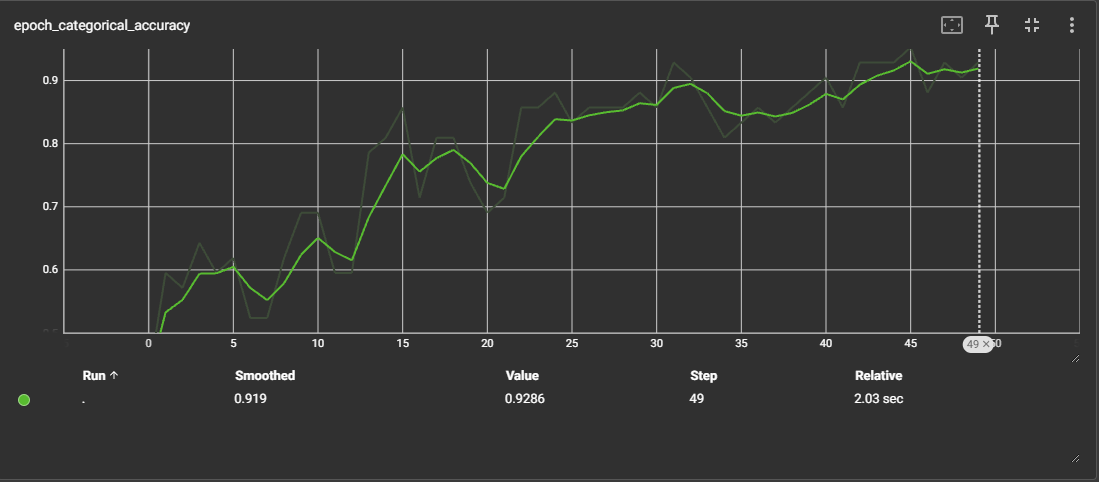}
    \caption{Categorical Accuracy of the LSTM model}
\end{figure}

\subsection{Analysis}
A confusion matrix will be used to evaluate all the models on the test data, where the true positives and true negatives are evaluated. For a model to have a good performance, the false positives and false negatives should be as low as possible.\\

The confusion matrix can only be used as a baseline for the model's accuracy. The final evaluation of the model will be measured on real-time footage. Adding noise or difficult footage to predict will truly show if the model can be used in real-world scenarios. From Figure 3 and Figure 4, it is evident that the model can predict a payment or evasion of passengers.\\ 

\subsection{Tools and Libraries}
All experiments will first be run on a Windows machine with a Ryzen 5 processor and a Nvidia RTX2070 Super GPU with 32GB RAM for all training purposes. The character action recognition LSTM model will be created with TensorFlow in the Python programming language.

In the demo of the project, the models will run on a laptop, with the openCV library the videos can be displayed for our model to predict payments and evasions caused by passengers on the bus. The end goal for future releases would be to run both models, simultaneously with at least one camera on a Raspberry Pi to predict fare evasions.

\vspace{12pt}
\section{Experimental Results}
To distinguish between these two possibilities, of payment or evasion, the LSTM model's confusion matrix on test data can show if the model has made accurate predictions. As shown in Figure 2, the LSTM model's predictions were accurate, with only having 2 false positive predictions.

\begin{figure}[H]
    \centering
    \includegraphics[width=8cm,height=7cm]{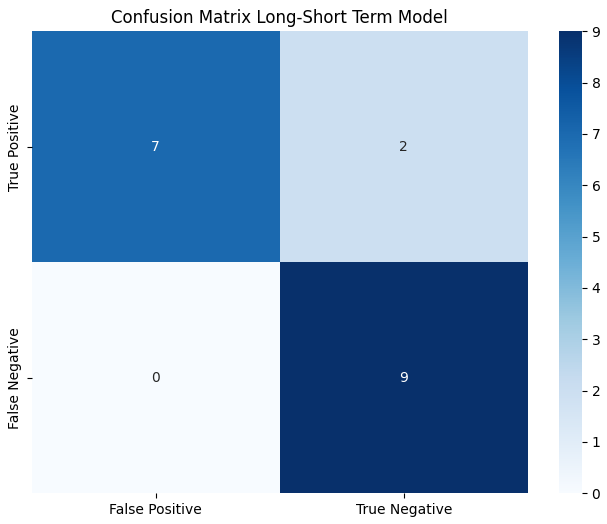}
    \caption{Confusion Matrix for Payments of the LSTM model}
\end{figure}
\vspace{16pt}
\begin{figure}[H]
    \centering
    \includegraphics[width=8cm,height=7cm]{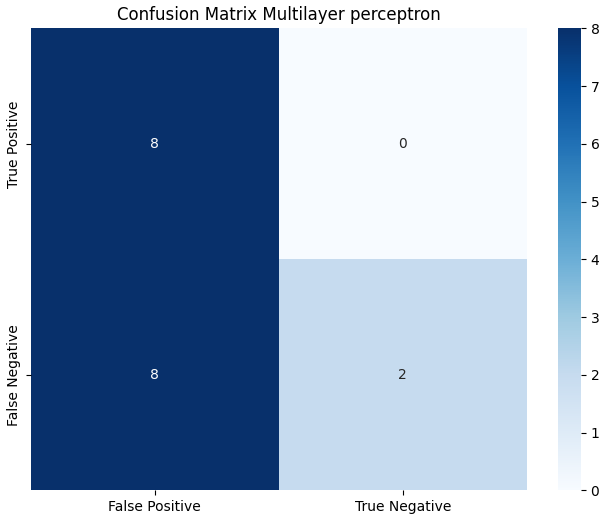}
    \caption{Confusion Matrix for Payments of the Multilayer Perceptron model}
\end{figure}
It can be derived from Figure 3 and Figure 4, that the LSTM model not only had a good confusion matrix on test data, but accurate predictions from the real-time footage as well.\\
\begin{figure}[H]
    \centering
    \includegraphics[width=7cm,height=6cm]{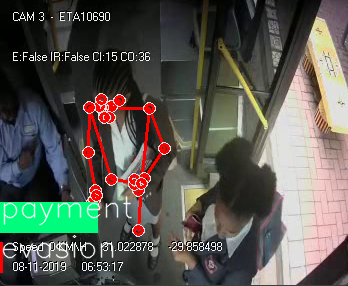}
    \caption{Payment Predicted with LSTM Model}
\end{figure}
\begin{figure}[H]
    \centering
    \includegraphics[width=7cm,height=6cm]{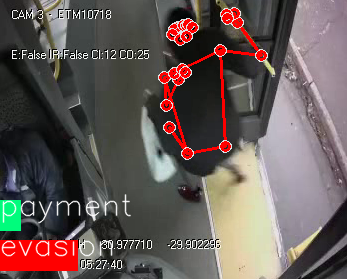}
    \caption{Evasion Predicted with LSTM Model}
\end{figure}
\vspace{16pt}

Changes in the LSTM and the MLP were compared using different hyperparameters. As seen in Figure 5, the MLP had suboptimal results compared to the LSTM model regarding its confusion matrix on test data. Upon inspection of Figure 5, the MLP model's predictions were quite inaccurate.\\

The multilayer perceptron had inadequate results on the test data, this is expected, considering it was developed from scratch and is inferior compared to the LSTM neural network that was created.

\vspace{12pt}
\section{Discussion}
The LSTM model in combination with a character re-identification (ReID) model, can be used to accurately detect the number of passengers committing fare evasions, which can reduce the number of passengers not paying for their fares. As seen in Figure 3 and Figure 4, the LSTM model can be a suitable solution for predicting if passengers paid for their bus fare.

An LSTM Deep Neural Network can be implemented to accurately predict if a passenger has committed fare evasion, which was the initial objective of the project.\\

We can see that in 2023, there has been a loss of one billion euros per year for 31 public transport companies, as mentioned in the literature \cite{FareEvasionInfoProvided}. With the help of modern CNN and LSTM models, it can be possible to prevent this from happening on such a significant scale.\\

The most obvious finding to emerge from the analysis is that human inspectors on the busses are not enough. Using sophisticated software such as an LSTM model, can help the amount of fare evasions to decrease.\\

An unexpected finding was the extent to which an LSTM model can predict passengers paying and not paying on buses. As they are usually used for time series predictions and not as a substitute for Convolution Neural Networks (CNN), to accurately predict the actions of humans with the help of keypoint extraction.\\

This finding broadly supports the work of other studies in this area, linking sophisticated models such as an LSTM model detecting actions of humans to classify if they are paying or not paying for their bus fare, with time series-based detection on tailgating passengers using human pose estimations \cite{FareEvasionTailgating}.\\

A possible explanation for an LSTM model to be a strong contender in predicting fare evasions is most likely the fact that keypoints of human actions in a series of frames are used over 3 seconds, which can be seen as a time-series problem that works on videos.\\

A note of caution is due here since the LSTM model was trained on human actions with the help of keypoint extractions, which can lead to inaccurate detections. The model was trained on people paying on the bus and not all types of payment methods for public transport.\\

These results provide further support for the hypothesis that a more sophisticated approach like an LSTM Neural Network can help detect if a passenger on the public bus services has paid for their fare or committed fare evasion.\\

These results provide some tentative initial evidence that with the help of Neural Networks, the number of fare evasions that occur in the public transport sector can be minimized. Increasing the number of inspectors only decreases this number by a small amount \cite{FareEvasionTimeSeries}. \\

\section{Conclusion}
The main goal of the current study is to determine if Deep Neural Networks like LSTM models can predict fare evasions for public transport companies. This study has identified ways to predict fare evasion with the use of an LSTM model by using keypoints over 3 seconds of passengers. The findings will be of interest to public transport companies that want to implement different ways to prevent or analyse which transport accumulates the most evasions.\\

Thirdly, the study did not include all possible ways for evasions to occur. It focused on public transport where passengers have to pay for their fares on the bus itself. Future studies should include other possible ways of passengers committing fare evasions and improve in ways to accurately predict evasions so that the public transport companies can analyse where the root of the problem is situated.

\vspace{12pt}
\section*{Acknowledgment}
I would like to thank Kuba South Africa for providing the footage of passengers for the model to be trained on and tested with.
\printbibliography
\end{document}